\documentclass[a4paper]{article}

\usepackage[margin=0.5in]{geometry}
\usepackage{amsmath}
\usepackage{bbold}
\usepackage{graphicx}
\usepackage{subcaption}
\usepackage{amsfonts}
\usepackage{booktabs}
\usepackage{xcolor}
\usepackage{enumitem}
\usepackage{authblk}
\usepackage{hyperref}
\usepackage[backend=bibtex,style=ieee,sorting=none,giveninits=true,maxbibnames=99,isbn=false,url=true,eprint=true,backref=false]{biblatex}
\renewbibmacro{in:}{}
\AtEveryBibitem{%
  \clearlist{language}%
}
\begin{document}
\title{A simple, strong baseline for building damage detection on the xBD dataset}
\author[1]{Sebastian Gerard}
\author[1,2]{Paul Borne-Pons}
\author[1]{Josephine Sullivan}
\affil[1]{KTH Royal Institute of Technology, Stockholm, Sweden\newline\texttt{\{sgerard,paulbp,sullivan\}@kth.se}}
\affil[2]{CentraleSupélec (Université Paris-Saclay), Paris, France \newline\texttt{paul.borne-pons@student-cs.fr}}
\date{\today}

\maketitle

\begin{abstract}
    We construct a strong baseline method for building damage detection by starting with the highly-engineered winning solution of the xView2 competition, and gradually stripping away components. This way, we obtain a much simpler method, while retaining adequate performance. We expect the simplified solution to be more widely and easily applicable. This expectation is based on the reduced complexity, as well as the fact that we choose hyperparameters based on simple heuristics, that transfer to other datasets. 
    We then re-arrange the xView2 dataset splits such that the test locations are not seen during training, contrary to the competition setup. In this setting, we find that both the complex and the simplified model fail to generalize to unseen locations. Analyzing the dataset indicates that this failure to generalize is not only a model-based problem, but that the difficulty might also be influenced by the unequal class distributions between events. \\\newline
    Code, including the baseline model: \url{https://github.com/PaulBorneP/Xview2_Strong_Baseline}
\end{abstract}

\section{Introduction}
\begin{figure}[ht]
    \centering
    \includegraphics[width=\textwidth]{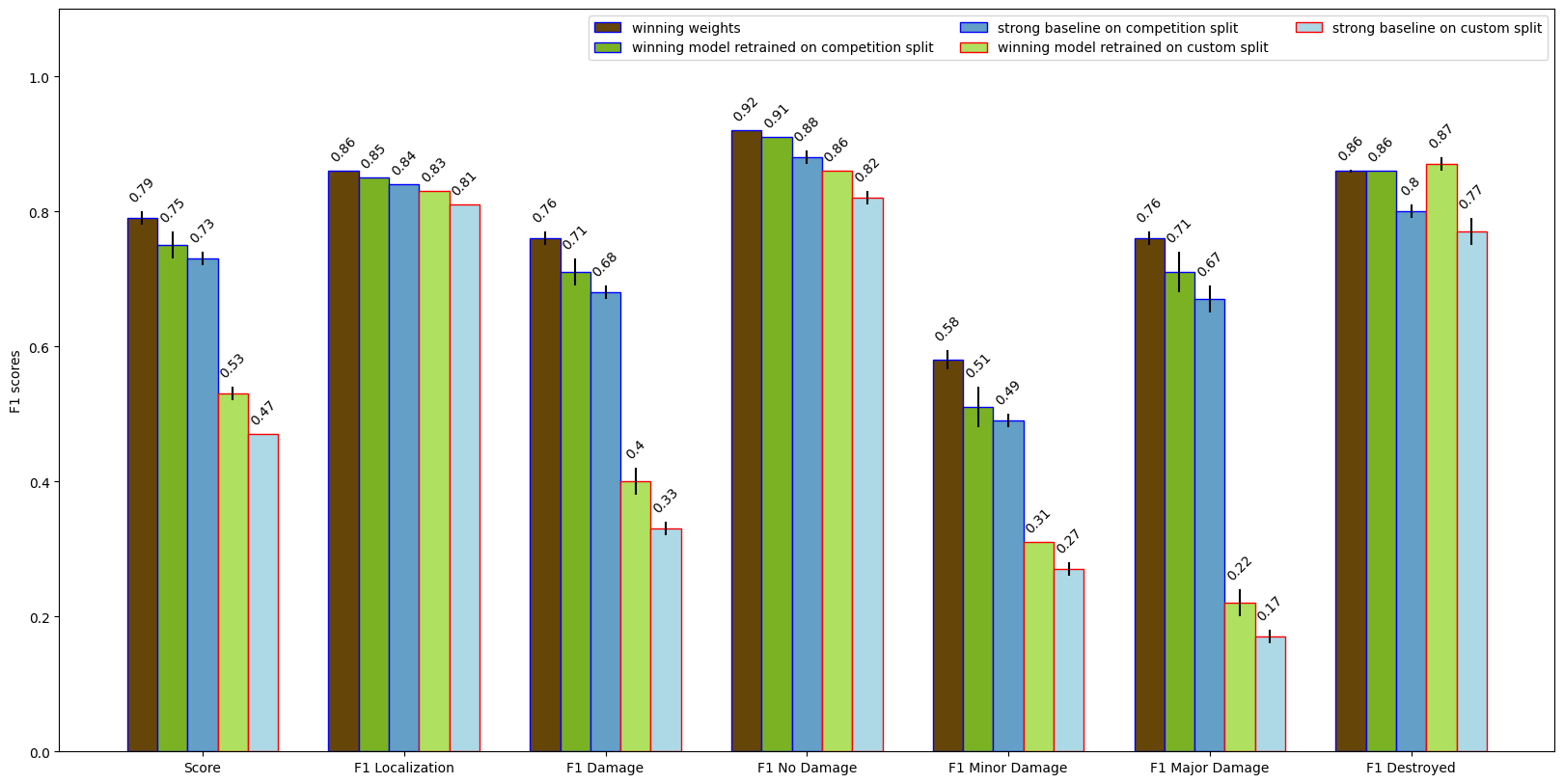}
    \caption{\textbf{Published weights, our reproduction, our simplified model, generalization.} We compare the following models: The  leftmost (brown) bar represents the published weights of the competition winning solution.~\cite{durnov_xview2_2020} The  $2^{nd}$ (dark green) bar uses the published code of the winning solution to retrain the model on our hardware. We can see a drop in performance that is not based on any intentional changes to the code. The  $3^{rd}$ (dark blue) bar shows our strong baseline model, derived from the winning solution by various simplification steps. It performs slightly worse than our reproduction. The two rightmost bar are, in order, the winning model (light green) and our strong baseline (light blue), retrained on a data split where the test disasters are not seen in training. Generalization proves difficult for both models. Although the strong baseline yields worse results, this difference is small, compared to the performance drop between the two splits. The drop is especially steep for 'minor damage' and 'major damage'. While the 'no damage' and 'destroyed' classes are easy to distinguish, it is difficult to clearly distinguish the two damage levels in between, so seeing the performance drop strongest in those two classes is not surprising. The results are based on individual ResNet34-U-Net models.}
    \label{fig:1}
\end{figure}

With progressing climate change, extreme weather events are projected to occur more often across many regions of the world.~\cite{change_ipcc_weather_2023} Since these events typically impact large areas, it is useful to incorporate large-scale observations from remote sensing systems in the disaster response. Modern computer vision methods can be used to process such large-scale observations automatically and provide various analysis results in the area of humanitarian assistance and disaster response (HADR).

As part of the \textsc{xView2} competition, the \textsc{xBD} building damage prediction dataset\cite{gupta_xbd_2019} was published. It contains high-resolution satellite image pairs from before and after various types of disasters, paired with high-quality building damage annotations for the post-disaster images. The natural starting point for research on this kind of problem now is the solution that won the \textsc{xView2} competition.~\cite{durnov_xview2_2020} However, since that solution was constructed with the goal of winning a competition, it is highly engineered and fine-tuned in many small ways, that make it hard to tweak and extend.

In this paper, we construct a simple, yet strong, baseline method by stripping away components of the \textsc{xView2}-winning solution step by step. It retains most of the performance of the original solution, while being much simpler, and therefore easier to use and extend in further research. We then show that both methods suffer greatly from generalization issues, when tested on a set of events of \textsc{xView2} that is not seen during training. This is contrary to the original setup of the \textsc{xView2} competition, in which images of each event are seen during training and testing.

We publish the simplified baseline as a LightningModule, as well as providing a LightningDataModule to load the \textsc{xBD} dataset itself. Both of these are part of the PyTorch Lightning framework~\cite{falcon_pytorch_2019}, that makes it easy to encapsulate all relevant steps for training and inference into one object, instead of having these details spread across various files and function calls, interspersed with the training logic.

\noindent In summary, our contributions are the following:

\begin{itemize}
\item We propose a strong baseline method based on the \textsc{xView2}-winning solution. We demonstrate that the step-by-step removal of various components only has a small influence on the performance of the method and the final method is only about 2 percentage points worse than our reproduction of the original method. 
  
\item We demonstrate that both the competition winner, as well as the strong baseline, suffers from a strong generalization failure in two of the four damage classes. This only becomes clear once we construct a dataset split, in which the events do not overlap between train and test set. 

\item We publish the strong baseline in an easily accessible form, to facilitate further research. 
\end{itemize}

\noindent\textbf{Structure} In \autoref{sec:dataset}, we first describe the xBD dataset, including class balance issues that might contribute to generalization problems. Then, we describe our strong baseline model in \autoref{sec:baseline}. In \autoref{sec:ablations}, we show the experimental results of simplifying the complex competition-winning model step by step until we arrive at our baseline model. This includes the description of the complexities of the winning solution. We move on to the question of generalization in \autoref{sec:generalization}, comparing the generalization performance of competition-winning model and the simplified model on a non-overlapping dataset split. \autoref{sec:generalization}.

\section{xBD: A Dataset for Assessing Building Damage from Satellite Imagery}
\label{sec:dataset}

\begin{figure}
    \centering
  \begin{subfigure}[b]{0.3\textwidth}
    \centering
    \includegraphics[width=\textwidth]{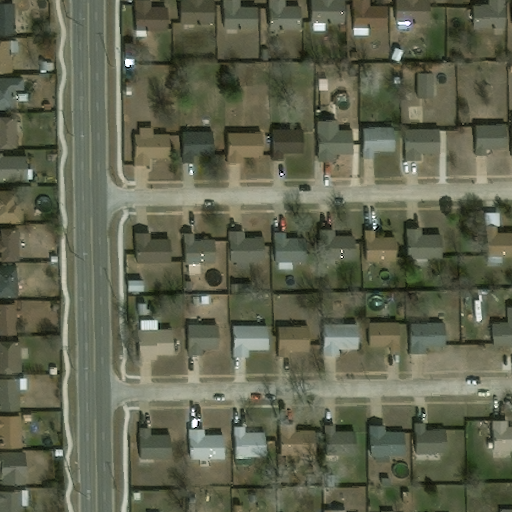}
    \caption{Pre-disaster}
  \end{subfigure}
  \hfill
  \begin{subfigure}[b]{0.3\textwidth}
    \centering
    \includegraphics[width=\textwidth]{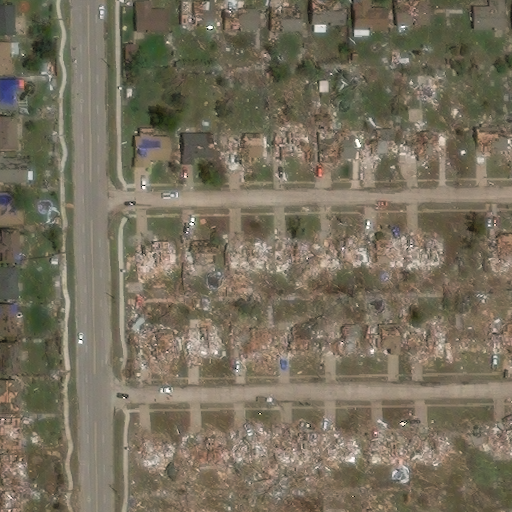}
    \caption{Post-disaster}
  \end{subfigure}
  \hfill
  \begin{subfigure}[b]{0.3\textwidth}
    \centering
    \includegraphics[width=\textwidth]{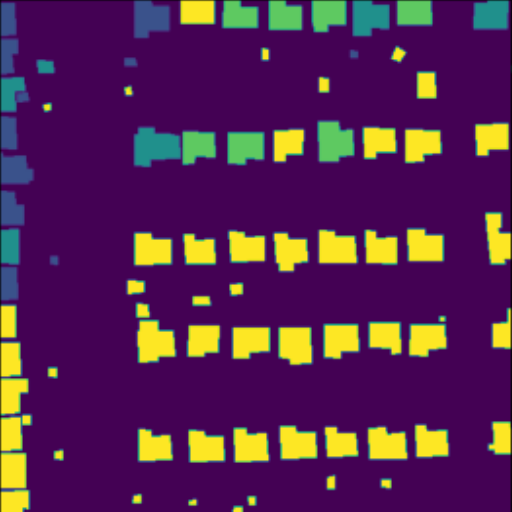}
    \caption{Building damage annotation}
  \end{subfigure}
    \caption{\textbf{xBD: Example images from the dataset} The images show a location impacted by the 2013 Moore tornado. The dataset contains one image from before and one after the disaster. The annotation indicates the location of buildings and the respective building damage class.}
    \label{fig:xbd-example}
\end{figure}

The \textsc{xBD} dataset~\cite{gupta_xbd_2019} is a satellite image dataset for building damage prediction. It contains images covering 22 disasters in 15 countries, including hurricanes, tornadoes, wildfires, earthquakes, floods and volcano outbreaks. It represents the task to detect buildings and classify them into one of four building damage classes, given one satellite image from before and one from after the disaster. The respective labels are high-quality annotations created by humans. Each building is consistently labelled with one class, even if only a part of the building is actually damaged. \autoref{fig:xbd-example} shows example images. The \textsc{xBD} dataset was the basis for the \textsc{xView2} challenge. For training in the ablation experiments, we combine the train subset (2799 image pairs) and the tier3 subset (6369 image pairs). In testing, we use the test subset, containing 933 image pairs from events that are already part of the training subsets. All images have a size of 1024² pixels.

\subsection{Generalization difficulty on the dataset-level}
\label{sec:generalization-dataset}

\begin{figure}[htpb]
    \centering
    \includegraphics[width=\textwidth]{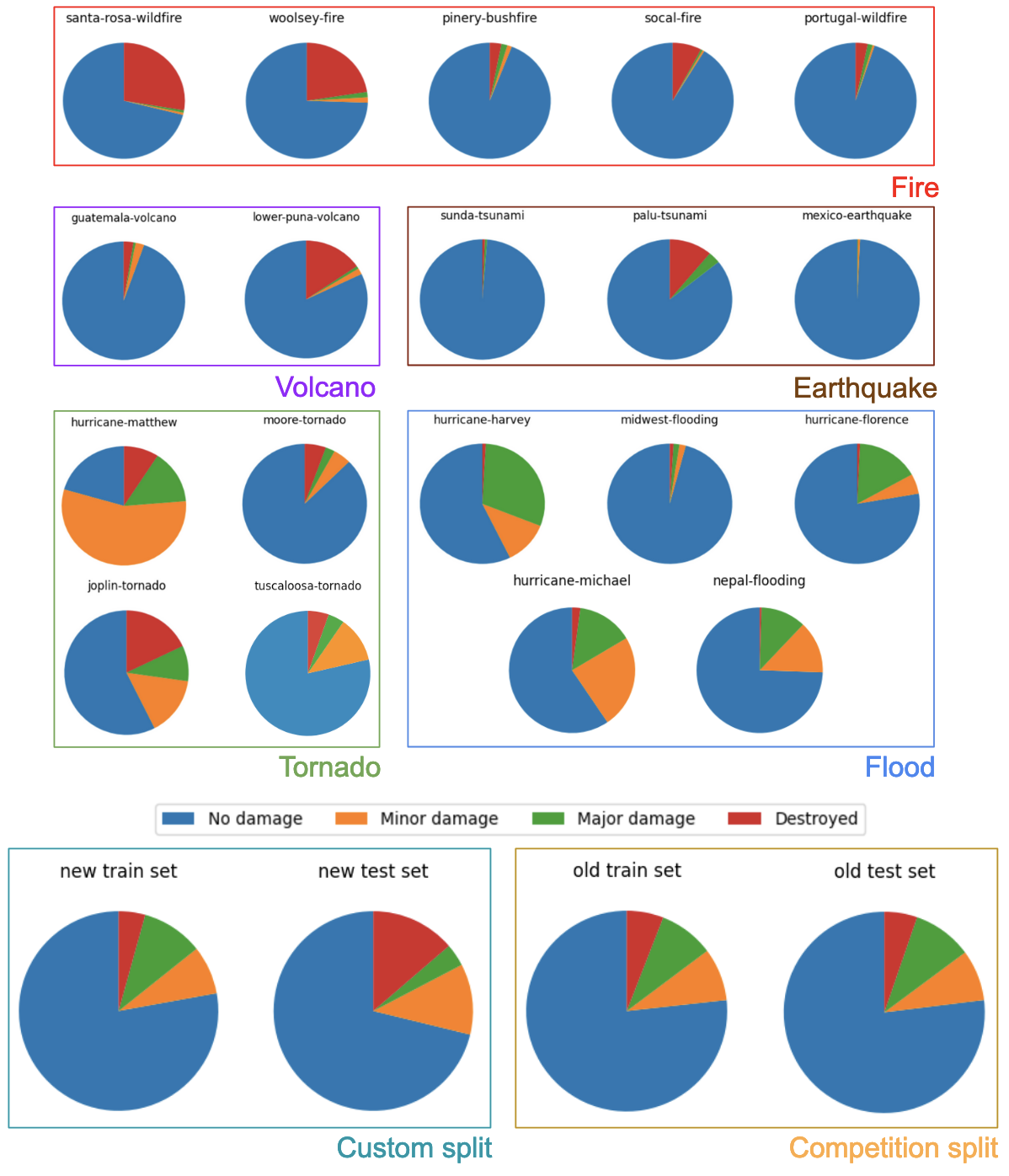}
    \caption{\textbf{Damage class distributions per event.} The distribution between events and event categories can differ greatly. For example, wildfire damages are mostly annotated as destroyed, while floods are mostly annotated as minor or major damage. This imbalance adds to the difficulty of training a model to perform well on all classes of all events.}
    \label{fig:class-pies}
\end{figure}

In this section we explore dataset properties that make it harder for the trained models to generalize well to unseen events. In \autoref{fig:class-pies}, we see how strongly the class distribution differs between events. For example, wildfire damages are rarely annotated as 'minor damage' or 'major damage', making it very difficult to learn to predict this class for this disaster type. While almost all distributions are dominated by the 'no damage' class, more than half of hurricane Matthew's building pixels are labeled as 'minor damage'. This is not the case for other hurricanes or tornadoes. This imbalance is based on the dataset composition and makes it hard for the model to perform equally well on all damage classes for all events, and also to generalize to events that might have very different class distributions. 

Part of the generalization problem is also that the way that damage classes look changes based on the geo-location. Minor damage looks different when it affects a large, expensive house in Asia than when it affects low-income housing in North America. This difficulty is inherent to the general task of building damage classification itself. Additionally, the distinction between damage classes from satellite images is no easy task in annotation either, especially given the different appearance of these classes across geographies and disaster types. While the 'no damage' and 'destroyed' classes are clearly distinguishable, the distinction between 'minor damage', 'major damage' and their respective neighboring classes is likely difficult, based only on the satellite images. Labeling with in-situ visitations could improve upon this, but this approach is of course completely impractical on the scale of this dataset. 

\subsection{Assessing generalization via a new dataset split}

\begin{figure}[ht]
    \centering
    \includegraphics[width=\linewidth]{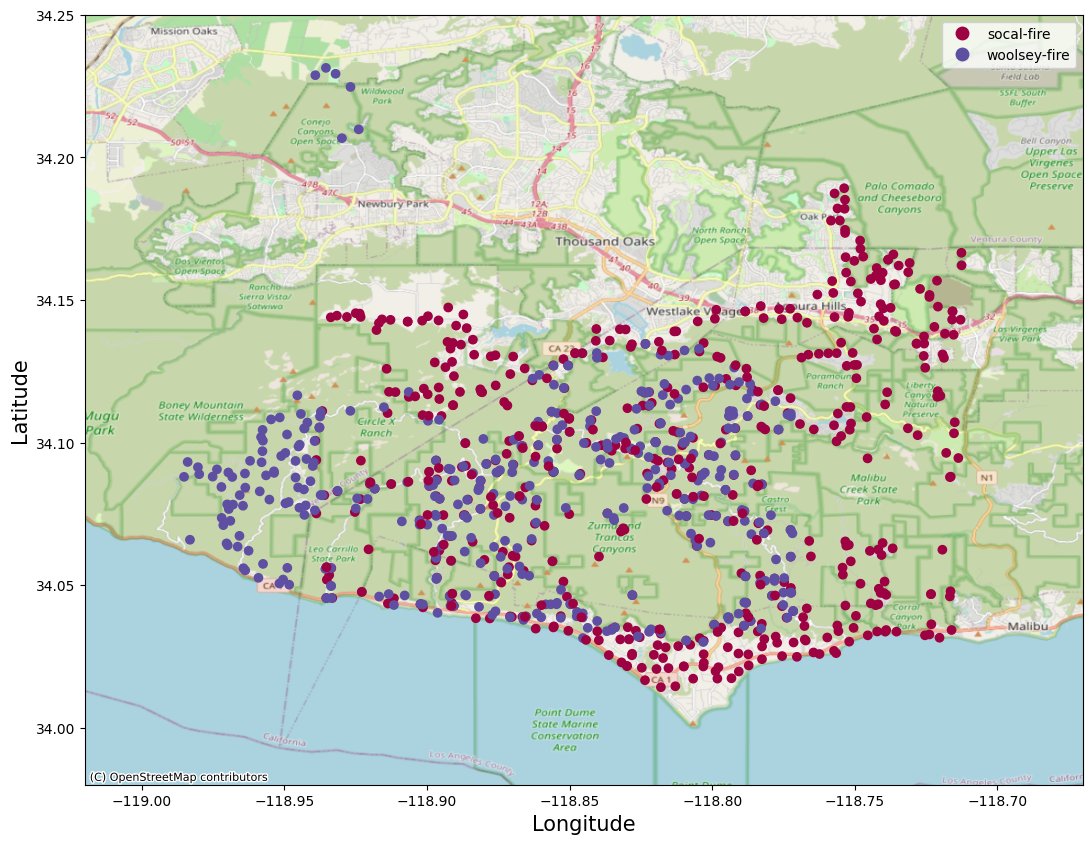}
    \caption{Spatial overlap between "socal fire" and "woolsey fire", two disasters that can be found in our dataset and that actually describe the same fire that occurred around  Los Angeles in November 2018. Map by \href{https://www.openstreetmap.org/copyright}{OpenStreetMap}.}
    \label{fig:spatial-overlap-1}
\end{figure}

\begin{figure}
    \centering
    \includegraphics[width=\linewidth]{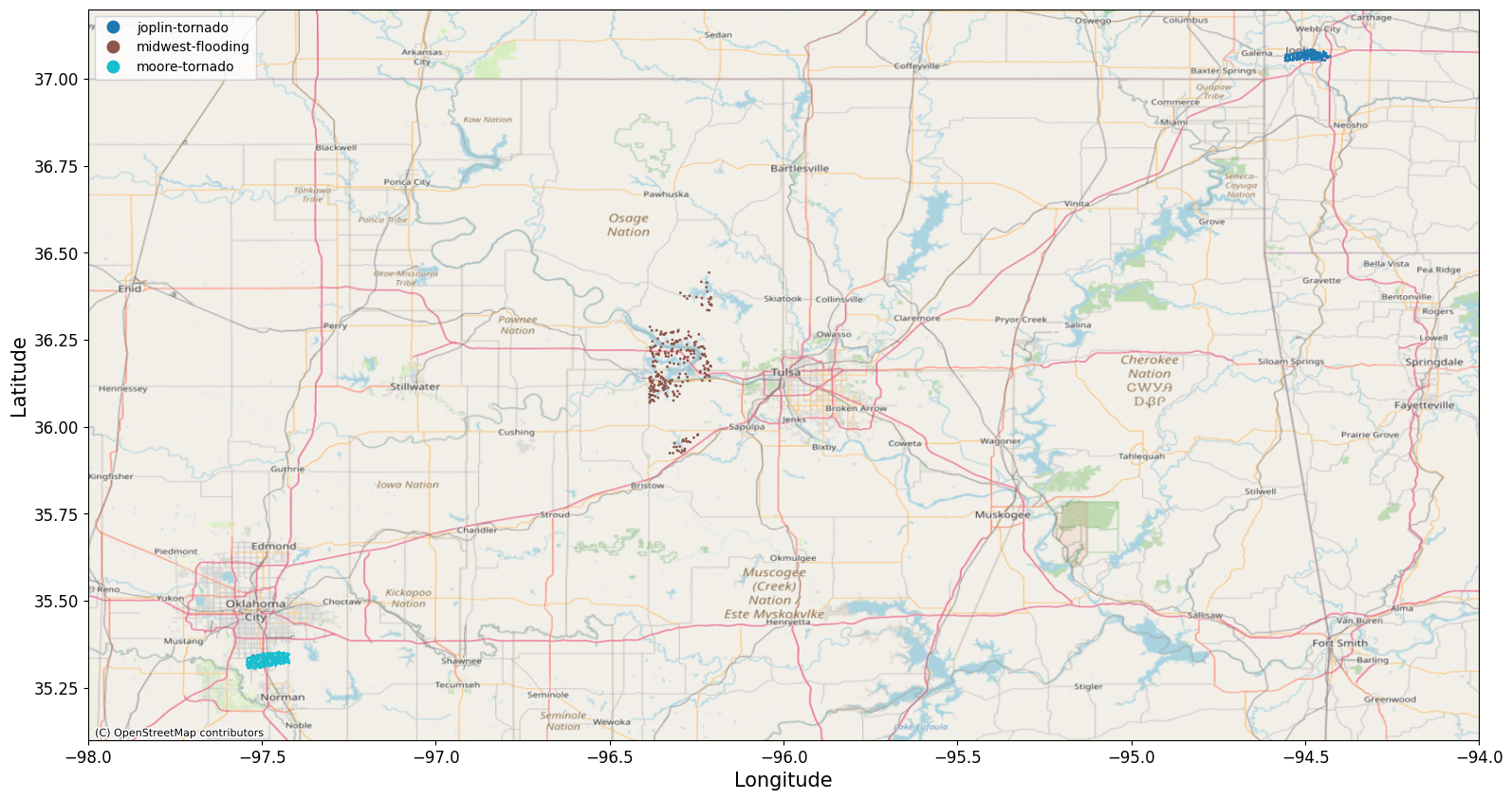}
    \caption{Spatial proximity between "joplin tornado", "moore tornado" and "midwestern floodings". Those disaster occurred in geographically close areas in the state of Oklahoma and were kept in the same set as prevention. Map by \href{https://www.openstreetmap.org/copyright}{OpenStreetMap}.}
    \label{fig:spatial-overlap-2}
\end{figure}

To be able to investigate how well models generalize, we want to test them on data that they have not seen during training. Since the original \textsc{xBD} dataset splits have images from all test events in the training set, we need to create a new split, that assigns each event exclusively to one of the subsets. At the same time, we want to retain at least one disaster event per event type in each of the subsets. For example, it does not make much sense to test on wildfires if the model has never seen wildfires during training. When plotting the images' locations on a map, we further found that several of the disaster events overlap spatially (see \autoref{fig:spatial-overlap-1}) or are in close spatial proximity (see \autoref{fig:spatial-overlap-2}). To ensure a fair generalization test, we keep all subgroups of spatially close events contained in the same subset. The new split is shown in \autoref{tab:new-data-splits}.


\begin{table}[ht]
    \centering
    \begin{tabular}{lcclc}
        \toprule
      \multicolumn{2}{c}{Train set} &  & \multicolumn{2}{c}{Test set} \\ \midrule
        event name & count &  & event name & count \\ 
        \cmidrule(r){1-2} \cmidrule(l){4-5} 

        lower-puna-volcano & 291 && tuscaloosa-tornado & 343 \\
        palu-tsunami & 155 & &guatemala-volcano & 23 \\
        mexico-earthquake & 159 && sunda-tsunami & 138 \\
        \textbf{socal-fire} & 1130 & &santa-rosa-wildfire & 300 \\
       \textbf{woolsey-fire} & 878 & &hurricane-matthew & 311 \\
        portugal-wildfire & 1869 & &--- & --- \\
        pinery-bushfire & 1845 & &--- & --- \\
        nepal-flooding & 619 & &--- & --- \\
        \textbf{midwest-flooding} & 359 & &--- & --- \\
        \textbf{moore-tornado} & 277 & &--- & --- \\
        \textbf{joplin-tornado} & 149 & &--- & --- \\
        hurricane-florence & 427 && --- & --- \\
        hurricane-harvey & 427 && --- & --- \\
        hurricane-michael & 441 && --- & --- \\
        \bottomrule
    \end{tabular}
\caption{\textbf{New dataset splits} To fairly evaluate the generalization performance of the models, we rearrange the dataset to be free of overlaps between the train and test set. This includes ensuring that disaster events that happened in close proximity are contained in the same data subset. Those disaster are in \textbf{bold} in the table}
\label{tab:new-data-splits}
\end{table}

\section{The strong baseline method}
\label{sec:baseline}

The strong baseline is a streamlined version of the \textsc{xView2} challenge winning solution~\cite{durnov_xview2_2020}, where we have kept the most important components of training and architecture. The ablation experiments for the intermediate steps between winning solution and simplified baseline are shown in \autoref{sec:ablations}. The strong baseline is built around an encoder-decoder network $f_{\text{\tiny enc-dec}}$ with skip connections and convolutional layers akin to a U-Net~\cite{ronneberger_u-net_2015}. $f_{\text{\tiny enc-dec}}$ is parameterized by $\Theta_1$. This network is applied independently to the before image $X_{\text{before}}$ and after image $X_{\text{after}}$ of the disaster site, each of which has width $W$, height $H$ and depth $D=3$: 
\begin{align}
  Z_{\text{before}} &= f_{\text{\tiny enc-dec}}(X_{\text{before}}, \Theta_1), &   Z_{\text{after}} &= f_{\text{\tiny enc-dec}}(X_{\text{after}}, \Theta_1)
\end{align}
with $Z_{\text{before}}, Z_{\text{after}} \in \mathbb{R}^{W \times H \times D_1}$.
These outputs are concatenated $Z_{\text{cat}} = [Z_{\text{before}}, Z_{\text{after}}]$ and input into a final small convolutional network $g$ consisting of $1 \times 1 \times 2D_1$ convolutions, parameterized by $\Theta_2$:
\begin{align}
  Z &= g(Z_{\text{cat}}, \Theta_2)
\end{align}
The final output is $Z \in \mathbb{R}^{W \times H \times \hat{C}}$, where $\hat{C}$ is the number of damage classes ($\hat{C}=4$: no damage, minor damage, major damage, destroyed). $Z$ contains the score for each pixel of belonging to each class. Each score in $Z$ is transformed to a scalar between 0 and 1 by applying the sigmoid function:
\begin{align}
  P &= \sigma(Z)
\end{align}
where $\sigma(\cdot)$ denotes the sigmoid function. It is applied to each entry of $Z$ independently and results in $P\in [0,1]^{W \times H \times \hat{C}}$. One can interpret the entry $P_{ij\hat{c}}$ of $P$ as the probability of pixel at location $(i,j)$ belonging to damage class $\hat{c}$.

Note: We represent the multi-class classification problem, where each pixel should ultimately be assigned to just one class, as multiple binary classification problems during training. While not strictly the correct approach (as each pixel can have only one label), this modelling has been shown~\cite{beyer_are_2020, kornblith_why_2021} to lead to better accuracy for supervised image classification.

\subsection{Building localization}

To discriminate building pixels from non-building pixels, we use a localization mask $L \in  \{0, 1\}^{W \times H}$. We compute the final segmentation mask $M \in \{0,1,2,3,4\}^{W \times H}$ as the output of our method by masking the damage segmentation with the localization mask. Class $0$ indicates the background class, while classes $1-4$ indicate the different damage classes: 

\begin{align}
M=(1+argmax(P_{ij\hat{c}}))\mathbb{1}_{L_{ij}=1}.
\end{align}

The localization mask $L$ can be computed in a number of ways, all of which yielded similar localization performance in our ablation studies:
\begin{enumerate}
\item  In the winning solution, the localization mask $L$ is created by thresholding the prediction output $P_{loc}$ of a separate localization model, that has the same architecture as the damage classification model. 
\item Our strong baseline predicts an additional class that distinguishes between building and no-building, independently of the damage classification.
\item Another approach we used on the way towards the strong baseline makes use of the building damage predictions that are already part of the model. The localization map can be created from the damage predictions by thresholding them (we use OTSU's method~\cite{otsu_threshold_1979}). If at least one of the binary classifications indicates that any level of building damage is present, it means there is also a building at this position. This is possible, because we formulated the problem as several independent binary classification problems, instead of using a softmax function. 
We used this method for the early versions of our solution, since a bug in the code led us to believe that the previously mentioned method was not working well enough.
\end{enumerate}

\subsection{Loss function}
\label{sec:loss-function}

The final version of our baseline predicts $C=5$ channels: four to predict building damage and one to predict the presence or absence of buildings. We keep the same notations as above but with $C$ instead of $\hat{C}$.

Learning the parameters, $\Theta = (\Theta_1, \Theta_2)$, of our complete network from labelled training data requires optimizing a loss function w.r.t. $\Theta$, given a training dataset. The loss in the winning solution is a combination of several standard loss functions used in semantic segmentation. Here, we review the mathematical definition of these losses and in turn also introduce our notation.

Let $Y = (Y_1, Y_2, \ldots, Y_C)$, where each $Y_c  \in \{0, 1\}^{W \times H}$, for $1 \leq c \leq C$, is the 2d binary matrix slice through the ground truth $Y$, representing which pixel locations have label $c$. Similarly let $P = (P_1, P_2, \ldots, P_C)$ be the respective predictions. 

For easier definition of the focal loss function, let  $\tilde{P}_c$ represent the predicted probability for the presence or absence of class $c$ at each pixel location as indicated by $Y_c$:
\begin{align}
  \tilde{P}_c &= (1-Y_c) \odot (1 - P_c) + Y_c \odot P_c
\end{align}
where $\odot$ represents element-wise multiplication. 
For a ground truth segmentation map $Y_c$ and a predicted segmentation map $P_c$ (and thus in turn also $\tilde{P}_c$), the standard loss functions used to train the strong baseline network for one input are:
\begin{align}
  L_{\text{\tiny focal}}(P_c, Y_c) &= -\frac{1}{W H} \sum_{i,j} (1 - \tilde{P}_{ijc})^{\gamma} \,\log(\tilde{P}_{ijc}) & \text{\footnotesize \textbf{Focal loss}}\\
  L_{\text{\tiny dice}}(P_c, Y_c) &= 1 - \frac{2\sum_{i,j} Y_{ijc}\, P_{ijc}}{\sum_{i,j} Y_{ijc} + \sum_{i,j} P_{ijc}}  & \text{\footnotesize \textbf{Soft dice loss}}
\end{align}
where $\gamma \geq 0$. 
The \textbf{combination loss}, for non-negative scalars $w_{\text{\tiny bce}}, w_{\text{\tiny focal}},$ and $w_{\text{\tiny dice}}$, is then defined as:
\begin{align}
  L_{\text{\tiny combo}}(P_c, Y_c) = w_{\text{\tiny focal}} L_{\text{\tiny focal}}(P_c, Y_c) + w_{\text{\tiny dice}} L_{\text{\tiny dice}}(P_c, Y_c).
\end{align}
Finally, the \textbf{overall training loss} over all classes, for one labelled input image pair is:
\begin{align}
  L(P, Y) = \sum_{c=1}^C w_c \, L_{\text{\tiny combo}}(P_c, Y_c),
\end{align}
where each $w_c$ is a non-negative scalar. 

The strong baseline uses $\gamma=2$, $(w_{\text{\tiny focal}}, w_{\text{\tiny dice}}) = ( 1, 1)$ and $w_c= \frac{1}{f_c}$ for $c \in \{1,2,3,4,5\}$, where $f_c$ is the relative frequency of class $c$ in the whole training dataset. As channel 0 represents the presence of a building and not the presence of background, the different classes are relatively balanced (compared to other segmentation tasks). 

\subsection{Further implementation details}
\label{sec:augmentations}

We use a single ResNet-34-based U-Net, which is the smallest among the winning solution's architectures. During training, the following geometric data augmentations are applied to the whole images:
{
  \begin{itemize}
  \item horizontal flipping
  \item rotations of 0, 90, 180, 270 degrees;
  \item random rotations in the range $[-10^{\circ}, 10^{\circ}]$ + random global scaling in the range $[.9, 1.1]$

\end{itemize}
}
For each labelled example the same augmentation is applied to before and after images. Afterwards, random square patches of side length in the range $[529, 715]$ are cropped from the data-augmented images with a bias towards patches that cover the building pixels from less frequent damage classes (using inverse frequency) and resized to side length 608.

\section{Experimental setup}
\label{sec:experimental-setup}

The experiments are mainly based on the \textsc{xView2} competition winner's code~\cite{durnov_xview2_2020}. It uses Python and PyTorch~\cite{paszke_pytorch_2019} to implement the models and all of the training and evaluation code. We additionally use Weights\&Biases~\cite{biewald_experiment_2020} for tracking experiments. 

\textbf{Hyperparameters} For training, we used an AdamW optimizer, a learning rate of 2e-4 and weight decay of 1e-6. The learning rate was iteratively halved on epoch 5, 11, 17, 23, 29 and 33. Furthermore, we used half-precision and DataParallel training on three RTX 2080Ti GPUs with a global batch size of 14.

\textbf{Evaluation metric} The competition score is computed as a weighted average of the localization F1 score $F1_{loc}$ and the harmonic mean of all building damage F1 scores $\text{F1}_{\text{\tiny dmg}} = 0.3~\text{F1}_{\text{\tiny loc}} + 0.7~\text{F1}_{\text{\tiny dmg}}$. Since localization and damage estimation are evaluated separately, the damage predictions are only evaluated where the ground truth labels show a building. Otherwise, localization errors would influence the damage prediction score. 

\textbf{Dataset splits} For the ablation studies in \autoref{sec:ablations}, we use the train and tier3 subsets as training data, and randomly split off 10\% into a validation set. The split is stratified over the disaster events, which is different from the completely random split used in the winning solution. To compute the test performance, we use the test set. The holdout set is not used. For the generalization experiments, we use the non-overlapping splits described in \autoref{sec:generalization-dataset}. The validation dataset is split off in the same way as in the ablation experiments. 

\section{Ablations: The path to a strong, but simple, baseline}
\label{sec:ablations}

\begin{table}[ht]
  \caption{Performance of the winning solution for the \textsc{xView2} challenge}
  \centering
\begin{tabular}{lrrrcrrrr}
  \toprule
  & & \multicolumn{2}{c}{\small \textbf{$F_1$ on main tasks}} & & \multicolumn{4}{c}{\small \textbf{$F_1$ score on damage classes}}\\
    \cmidrule(ll){3-4} \cmidrule(ll){6-9}
    \textbf{Model} &  \textbf{Score} &  Local. & Damage & & None & Minor & Major & Destroyed\\
    \midrule
  Ensemble (4 architectures à 3 seeds) & 0.804 & 0.862 & 0.779 & & 0.929 & 0.615 & 0.778 & 0.872\\
  ResNet-34 (3 seeds) & 0.792 & 0.859 &0.764 & & 0.921 & 0.592 & 0.765 & 0.866 \\  
  ResNet-34 (best seed) & 0.789 & 0.858 & 0.759& & 0.916 & 0.583 & 0.769 & 0.862 \\  
  ResNet-34 (best seed, our rerun) & 0.770 & 0.854 & 0.734& & 0.911 & 0.544 & 0.741 & 0.860 \\  
  
  \bottomrule
\end{tabular}
\label{tab:ensemble-vs-single}
\end{table}

\begin{table}[]
\small
\caption{\textbf{Ablation studies:} We start from the re-run of the \textsc{xView2} winner's solution (top row), and reduce its complexity step by step. We reach a much simpler baseline approach, while the \textsc{xView2} competition score only drops from 0.75 to 0.73. Our model does not use a separate finetuning step with a jump back up in learning rate. We use a single model for localization and classification, instead of two separate ones. This includes not validating the model on the competition metric, while using the localization model, but on the loss function instead. We use an equal-weighted combination of Focal and soft Dice loss. We use class weights for the losses and crop-selection that reflect the class distribution, instead of optimized weights. We only use geometric augmentations. We do not dilate building classes in training, nor the 'minor damage' class in testing. The competition scores are the mean and standard deviation over three repetitions with varying random seeds.\\}
\begin{tabular}{@{}cccccccccc@{}}
\toprule
\begin{tabular}[c]{@{}c@{}}Classif. \\ pretraining\end{tabular} &
  \begin{tabular}[c]{@{}c@{}}Separate \\ Finetuning\end{tabular} &
  \begin{tabular}[c]{@{}c@{}}Loc\\ method\end{tabular} &
  \begin{tabular}[c]{@{}c@{}}Tuned weights \\in combo loss \\function\end{tabular} &
  \begin{tabular}[c]{@{}c@{}}Optimized \\loss \\ class weights\end{tabular} &
  \begin{tabular}[c]{@{}c@{}}Photom.\\augs\end{tabular} &
  \multicolumn{2}{c}{\begin{tabular}[c]{@{}cc@{}}\multicolumn{2}{c}{Dilation}\\ train &    test\end{tabular}} & 
  \begin{tabular}[c]{@{}l@{}}Optimized \\cropping \\class weights\end{tabular} &
  \begin{tabular}[c]{@{}l@{}}Competition \\score\end{tabular} \\ \midrule
Localization & \checkmark & Loc model & \checkmark & \checkmark & \checkmark & \checkmark & \checkmark & \checkmark & 0.75 ± 0.02 \\
ImageNet          & \checkmark & Loc model &  \checkmark & \checkmark & \checkmark & \checkmark & \checkmark & \checkmark & 0.75 ± 0.01 \\
ImageNet          & -  &  Loc model &  \checkmark& \checkmark & \checkmark & \checkmark & \checkmark & \checkmark & 0.74 ± 0.0  \\
ImageNet          & -  & OTSU & \checkmark & \checkmark & \checkmark & \checkmark & -  & \checkmark & 0.76 ± 0.01 \\
ImageNet          & -  & OTSU & -  & -  & \checkmark & \checkmark & -  & \checkmark & 0.74 ± 0.01 \\
ImageNet          & -  & OTSU & -  & -  & -  & \checkmark & -  & \checkmark & 0.74 ± 0.00  \\
ImageNet          & -  & OTSU & -  & -  & -  & - & -    & \checkmark & 0.73 ± 0.01 \\
ImageNet          & -  & OTSU & -  & -  & -  &  - & -   & -  & 0.73 ± 0.01 \\ 
ImageNet          & -  & Loc channel & -  & -  & -  &  - & -   & -  &  0.73 ± 0.01 \\ \bottomrule
\end{tabular}
\label{tab:ablations}
\end{table}

We want to find out which components of the \textsc{xView2} competition's winning solution (\#1 solution) are important for its success, and which can be left out, without losing a lot of performance.

\textbf{Large ensemble} The \#1 solution consists of an ensemble of four architectures, each trained on three random seeds. We investigate the effect of this ensemble approach by using the published model weights~\cite{durnov_xview2_2020} of the original author. \autoref{tab:ensemble-vs-single} shows that if we simply pick the best performing seed of the smallest model, a ResNet-34-based U-Net, we only lose 1.5pp (percentage points) in the competition score. This way, we greatly reduce the cost of training and inference. In the following ablation experiments, we will retrain the model with various changes to the initial setting. 

To have a fair point of comparison, we re-run the training of the best individual model among the published models. This model's performance will be used as the upper bound of performance that we compare all ablated model versions to. Notably, the retrained model has 1.9pp lower performance than the published weights. One factor that might cause this is the change from randomly split train/val sets in the original code, to a stratified split, in our experiment. Another factor is that the original hyperparameters might be optimized for the used random seeds, which ensure reproducibility on the same machine, but lead to different behavior when moving to our hardware.

\textbf{Single model} Starting from the ResNet-34 U-net, we remove different aspects of the model, step by step. The results are shown in \autoref{tab:ablations}. The first row represents the re-run of the best individual model as a reference point. We will walk through the results here, to also explain the intricacies and tricks used to squeeze out every last bit of performance, and win the \textsc{xView2} competition. The resulting competition scores can be seen in \autoref{tab:ensemble-vs-single}, but won't be mentioned in the text, since the performance differences are all rather small.

\textbf{Pretraining \& finetuning} The original approach uses two separate models, one for binary building localization, one for damage classification. The localization model is trained first, and then used as a starting point for the classification model. We use an ImageNet-pretrained model as the starting point for classification instead. After training the classification model, the \#1 solution then performs something the original author calls tuning. This consists of training for 3 more epochs, with a learning rate that is halved every epoch and slightly changed augmentations. The learning rate is set to a value that's higher than the one at the end of the training, constituting a slight warm restart. We remove this separate finetuning step. We increase the number of training epochs from 20 to 40, to account for the fact that ImageNet pretraining is less close to the building damage segmentation than the localization pretraining. Additionally, we simplify the thresholding of the localization prediction, which in the \#1 solution is a combination of three different threshold conditions, applied to localization and building damage channels. We only retain the simplest threshold, which is a fixed threshold applied only to the localization prediction.

\textbf{Integrating localization and classification} As a next step, we remove the separate localization model. Instead, we use the building damage predictions, that are modeled as four independent binary classification problems. If any of them indicates the presence of building damage, the respective pixel must contain a building. We apply OTSU-thresholding~\cite{otsu_threshold_1979} to each building damage channel and combine them with logical OR operations, to compute the localization mask. At the same time, we also remove test-time dilation, which dilates 'minor damage' predictions to overwrite 'no damage' predictions as a post-processing step.

\textbf{Tuned class and loss weights} The \#1 solution uses a weighted combination of multiple loss functions (see \autoref{sec:loss-function}). For the ResNet34-based model, these are focal and soft dice loss. For other architectures, they also mix in a cross-entropy loss. We replace the weights for the loss functions with equal weights for both losses. Furthermore, each per-class loss is weighted in a way that is likely optimized via search and/or hand-tuned. We replace these weights with the inverse relative frequency of each class in the training set. The original class weights do not quite follow this distribution, and also overemphasize the difficult 'minor damage' class.

\textbf{Training-time augmentations} The \#1 solution uses both geometric augmentations (see \autoref{sec:augmentations}), as well as a range of photometric augmentations: Channel-wise addition in RGB and HSV space, CLAHE, Gaussian noise, blurring, and changes in saturation, brightness and contrast.  We remove the photometric augmentations and retain only the geometric ones. The \#1 solution also dilates all building damage classes during training, again preferring the 'minor damage' class if multiple dilated classes overlap. We remove this training-time dilation. During random cropping, the \#1 solution considers multiple candidate crops and computes a preference score, again based on per-class weights multiplied with the number of pixels of each class in the candidate crop. The crop with the highest preference score is used. Oversampling buildings makes sense, so we keep this aspect, but again replace the per-class weights with distribution-based weights.

\textbf{Localization as fifth channel} Lastly, we replace the localization via OTSU thresholding with simply predicting the presence of buildings as a separate channel. This had been our first choice, since it is simple and clean, but due to a problem in the code, it seemed like this option was not working well. Instead of the adaptive OTSU threshold, we also go back to the simple fixed threshold value set in the original solution. Both options should work in practice. We make this last change and receive a much simpler method than the one we started with, while only losing two percentage points in competition score across all of the changes made.

\subsection{Aspects we kept}

During the ablation process, we attempted some modifications that resulted in a severe drop in performance. Notably, when we removed the dice part of the combo loss. The dice part in the loss appears to be necessary for learning good building localization.

Another critical aspect was the importance of choosing the right batch size during training. A batch size that is too small can negatively impact the stability of training and ultimately affect the final score. This issue becomes particularly problematic when using a weighted loss function, as some classes may be underrepresented in the dataset. Consequently, computing the loss for a few images can lead to encountering edge cases, hence making the training unstable.

Other aspects that we did not try to change were the initial learning rate, learning rate schedule and optimizer, as well as the fixed threshold used for the localization prediction. The building damage predictions are aggregated via argmax. We also did not try to change the relatively standard geometric data augmentations.

\section{Generalization: difficulties on the model-level}
\label{sec:generalization}

To assess the generalization performance of both models, and whether the simpler model is maybe less overfit to the dataset, we train the model on the new data splits shown in \autoref{tab:new-data-splits}. We use the same hyperparameters as in \autoref{sec:ablations}, except that the class weights have been recomputed for the new training set. The results are shown in the respective two rightmost bars in the columns of \autoref{fig:1}.

We see that both models overall perform much worse than on the original data split. The localization, as well as the detection of the 'no damage' and  'destroyed' class are only a few percentage points worse than on the old split, confirming that they are distinct enough to be easily distinguishable. However, the performance on the 'minor damage' class drops roughly 20\% and the 'major damage' class even drops roughly 50\% for both models. These large gaps in performance between the old, overlapping split, and the new, non-overlapping split, demonstrates a big problem of generalization. Notably, the 'minor damage' class is the one that received special attention in the \textsc{xView2} winner's solution. For example, it received a disproportionately high class weight in training, and predictions on it were artificially dilated as a post-processing step. This can explain some of the higher performance of the winning solution. 
Comparing the simplified baseline with the winning solution, the winning solution is again clearly better than the simplified model, but the differences between the models' performances is much smaller than the generalization gap.

Based on the observations about class distributions differing between events (see \autoref{sec:generalization-dataset}), we also investigate if the performance on the individual events in the test set are associated with the distance between either the overall training set's class distribution or the specific event type's class distribution in training. In both cases, we do not find a strong association. The detailed results are not reported here.

\section{Conclusion}
\label{sec:conclusion}

We have shown that a greatly simplified version of the \textsc{xView2} competition winner's approach still achieves an adequate performance. In the simplified approach, various hyperparameter choices are made in clear, principled ways, making it easy to train the model on different datasets or different splits of the same dataset. Other parts of the complex approach are removed without losing much performance. By retraining both models on a non-overlapping dataset split, we find that generalization to unseen locations is a problem for both models. Due to the simplified version, we know that this problem is not caused by the winning model being overfit to the original dataset. After showing that the class distributions vary greatly between events, we assume that this imbalance in the dataset composition likely contributes to the difficulty of generalizing to unseen events.

\printbibliography

\newpage
\appendix

\section{Contributions according to the CRediT system}

Description of the CRediT system: \url{https://credit.niso.org/}

\begin{itemize}
    \item Sebastian Gerard: Conceptualization, Formal analysis, Investigation, Methodology, Software, Writing – original draft, Writing – review \& editing

    \item Paul Borne-Pons: Formal analysis, Investigation, Methodology, Software, Validation, Visualization, Writing – original draft

    \item Josephine Sullivan: Conceptualization, Funding acquisition, Project administration, Resources, Supervision, Writing – review \& editing

\end{itemize}

\clearpage
\newpage

\iffalse

\fi

\end{document}